# A Topic Model for Melodic Sequences

**Athina Spiliopoulou** A.SPILIOPOULOU@ED.AC.UK
**Amos Storkey** A.STORKEY@ED.AC.UK
School of Informatics, University of Edinburgh

## Abstract

We examine the problem of learning a probabilistic model for melody directly from musical sequences belonging to the same genre. This is a challenging task as one needs to capture not only the rich temporal structure evident in music, but also the complex statistical dependencies among different music components. To address this problem we introduce the Variable-gram Topic Model, which couples the latent topic formalism with a systematic model for contextual information. We evaluate the model on next-step prediction. Additionally, we present a novel way of model evaluation, where we directly compare model samples with data sequences using the Maximum Mean Discrepancy of string kernels, to assess how close is the model distribution to the data distribution. We show that the model has the highest performance under both evaluation measures when compared to LDA, the Topic Bigram and related non-topic models.

## 1. Introduction

Modelling the real-world complexity of music is an interesting problem for machine learning. In Western music, pieces are typically composed according to a system of musical organization, rendering musical structure as one of the fundamentals of music. Nevertheless, characterizing this structure is particularly difficult, as it depends not only on the realization of several musical elements, such as scale, rhythm and meter, but also on the relation of these elements both within single time frames and across time. This results in an infinite number of possible variations, even within pieces from the same musical genre, which are typically built according to a single musical form.

To tackle the problem of melody modelling we propose the Variable-gram Topic model which employs a Dirichlet Variable-Length Markov Model (Dirichlet-VMM) (Spiliopoulou & Storkey, 2011) for the parametrisation of the topic distributions over words. The Dirichlet-VMM models the temporal structure by learning contexts of variable length that are indicative of the future. At the same time, the latent topics represent different music regimes, thus allowing us to model the different styles, tonalities and dynamics that occur in music. The model does not make any assumptions explicit to music, but it is particularly suitable in the music context, as it is able to model temporal dependencies of considerable complexity without enforcing a stationarity assumption for the data. Each sequence is modelled as a mixture of latent components (topics), and each component models Markov dependencies of different order according to the statistics of the data that are assigned to it.

To evaluate the performance of the model we perform a comparative analysis with related models, using two metrics. The first one is the average next-step prediction log-likelihood of test sequences under each model. The second is the Maximum Mean Discrepancy (MMD) (Gretton et al., 2006) of string kernels computed between model samples and test-data sequences. In both evaluations, we find that using topics improves performance, but it does not overcome the need for a systematic temporal model. The Variable-gram topic model, which couples these two strategies has the highest performance under both evaluation objectives.

The contributions of this paper are: (a) We introduce the Variable-gram Topic model, which extends the topic modelling methodology by considering conditional distributions that model contextual information of considerable complexity. (b) We introduce a novel way of evaluating generative models for discrete data. This employs the MMD of string kernels to directly compare model samples with data sequences.

## 2. Background

A number of machine learning and statistical approaches have been suggested for music related prob-

---

lems. Here we discuss methods that take as input discrete music sequences and attempt to model the melodic structure. Lavrenko & Pickens (2003) propose Markov Random Fields (MRFs) for modelling polyphonic music. The model is very general, but in order to remain tractable much information is discarded, thus making it less suitable for realistic music. Weiland et al. (2005) propose a Hierarchical Hidden Markov Model (HHMM) for pitch. The model has three internal states that are predefined according to the structure of the music genre examined. Eck & Lapalme (2008) propose an LSTM Recurrent Neural Network for modelling melody. The network is conditioned on the chord and certain previous time-steps, chosen according to the metrical boundaries. Paiement et al. (2009) provide an interesting approach that incorporates musical knowldege in the melody modelling task. They define a graphical model for melodies given chords, rhythms and a sequence of Narmour features, which are extracted from an Input-Output HMM conditioned on the rhythm.

A very successful line of research examines the transfer of methodologies from the fields of statistical language modelling and text compression to the modelling of music. Dubnov et al. (2003) propose two dictionary-based prediction methods, Incremental Parsing (IP) and Prediction Suffix Trees (PSTs), for modelling melodies with a Variable-Length Markov model (VMM). Despite its fairly simple nature the VMM is able to capture both large and small order Markov dependencies and achieves impressive musical generations. Begleiter et al. (2004) study six different alogrithms for training a VMM. These differ in the way they handle the counting of occurences, the smoothing of unobserved events and the variable-length modelling. Spiliopoulou & Storkey (2011) propose a Bayesian formulation of the VMM, the Dirichlet-VMM, for the problem of melody modelling. The model is shown to significantly outperform a VMM trained using the PST algorithm. Finally, an interesting application of dictionary-based predictors in the music context is presented in Pearce & Wiggins (2004). They describe a multiple viewpoint system comprising a cross-product of Prediction by Partial Match (PPM) models.

# 3. The Variable-gram Topic Model

In this section we introduce the Variable-Gram Topic model, which we later apply to melodic sequences. In the context of music modelling, documents correspond to music pieces and words correspond to notes. The Variable-Gram Topic model extends Latent Dirichlet Allocation (LDA) by employing the Dirichlet Variable-Length Markov model (Dirichlet-VMM) (Spiliopoulou & Storkey, 2011) for the parametrisation of the topic

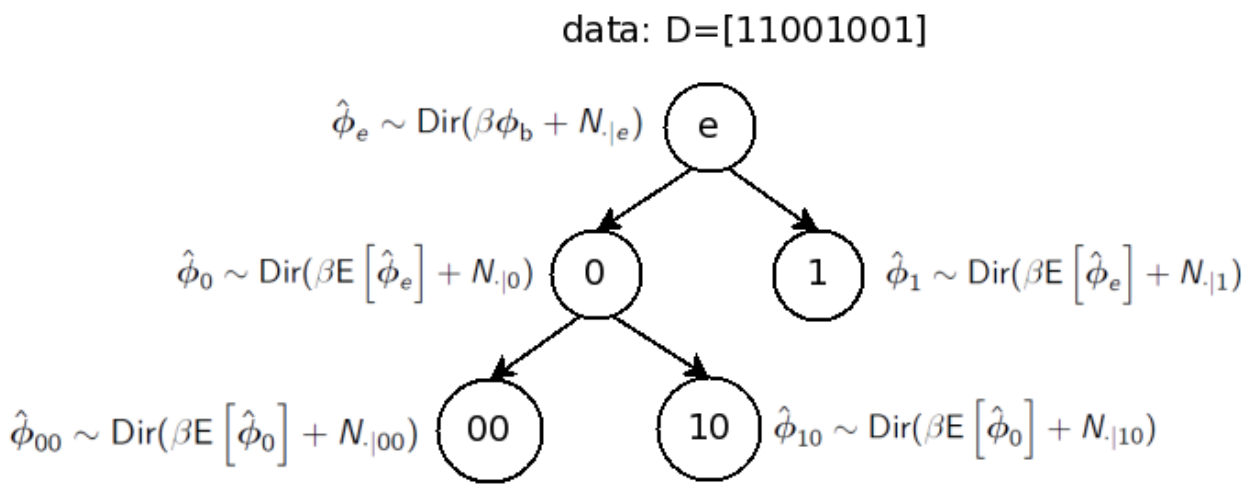

*Figure 1.* An example Dirichlet-VMM tree for a binary sequence. Contexts 01 and 11 are only observed once and thus are not included in the tree. Note that for readability, contexts in this figure are denoted in chronological order.

distributions over words. We begin with a description of the Dirichlet-VMM.

## 3.1. The Dirichlet-VMM

The Dirichlet-VMM is a Bayesian hierarchical model for discrete sequential data defined over a finite alphabet. It models the conditional probability distribution of the next symbol given a context, where the length of the context varies according to what we actually observe. Long contexts that occur frequently in the data are used during prediction, while for infrequent ones, their shorter counterparts are used.

Similarly to a VMM, the model is represented by a suffix tree that stores contexts as paths starting at the root node; the deeper a node in the tree the longer the corresponding context. The depth of the tree is upper bounded by $L$, the maximum allowed length for a context. The tree is not complete; only contexts that occur frequently enough in the data and convey useful information for predicting the next symbol are stored. The Probabilistic Suffix Tree algorithm for constructing a VMM tree is detailed in Ron et al. (1994).

In contrast to the VMM, parameter estimation in the Dirichlet-VMM is driven by Bayesian inference. Let $w$ denote a symbol from the alphabet and $j$ index the nodes in the tree, with $\mathbf{c}_j = w_1 \dots w_\ell, \ell \in \{1, \dots, L\}$, denoting the context of node $j$. Each node $j$ is identified by the conditional probability distribution of the next symbol given context $\mathbf{c}_j$, which we denote by $\phi_{i|j} \equiv P(w = i|\mathbf{c}_j)$. In the Dirichlet-VMM this distribution is modelled through a Dirichlet prior centred at the parent node, $\boldsymbol{\phi}_j \sim \mathrm{Dirichlet}(\beta\boldsymbol{\phi}_{\mathbf{pa}(j)})$, where $\beta$ denotes the concentration parameter of the Dirichlet distribution, $\mathbf{pa}(j)$ denotes the parent of node $j$, with corresponding context $\mathbf{c}_{\mathbf{pa}(j)} = w_1 \dots w_{\ell-1}$, and we have used the bold notation $\boldsymbol{\phi}_j$ to denote the parameter vector $\boldsymbol{\phi}_{\cdot|j}$. An example Dirichlet-VMM is depicted in Figure 1.

Due to the conjugacy of the Dirichlet distribution to the multinomial, posterior inference in this model is exact. Let $\hat{\phi}_{i|j} \equiv P(w = i|\mathbf{c}_j, D)$ denote the es-timate for $\phi_{i|j}$ after observing data $D$. We have $\hat{\boldsymbol{\phi}}_j \sim \text{Dirichlet}(\beta \text{E}[\hat{\boldsymbol{\phi}}_{\mathbf{pa}(j)}] + N_{\cdot|j})$, where $N_{\cdot|j}$ denotes the counts associated with context $j$ in the data and $\text{E}[\cdot]$ denotes expectation.

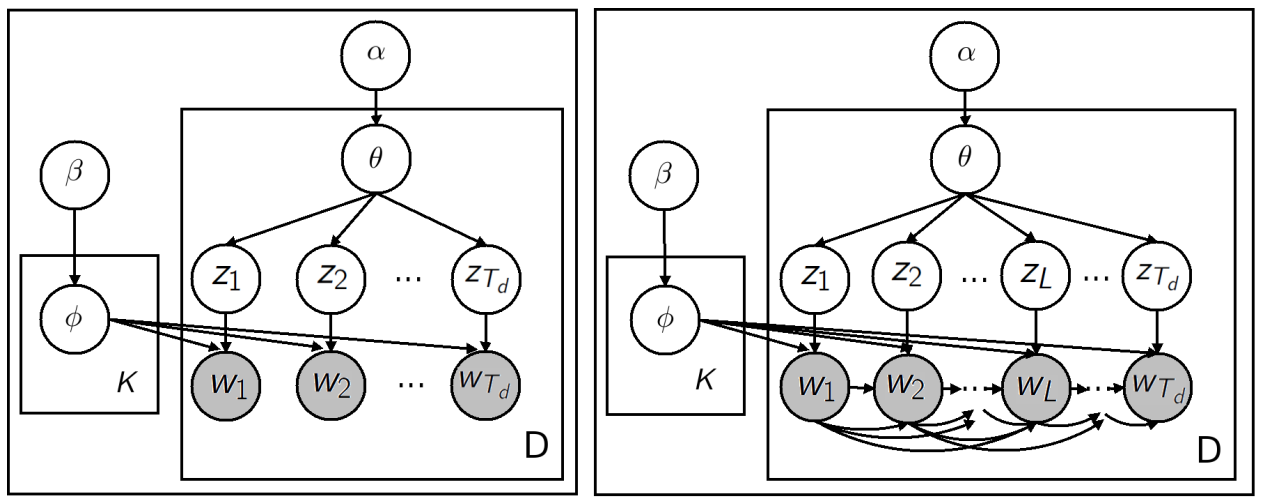

(a) (b)

*Figure 2.* Graphical models: (a) LDA without the plate notation for words (b) Variable-gram Topic model.

During prediction only the leaf nodes are used. Given an observed context, we start at the root node and follow the path labelled by successively older symbols from the context, until we reach a leaf node, from which we read the predictive distribution. The hierarchical construction of the Dirichlet-VMM allows us to maintain information coming from the shorter contexts into the predictive probabilities of the longer ones. It is related to the hierarchical Dirichlet language model (Mackay & Peto, 1995), where instead of a single fixed order we now consider n-gram statistics of variable order, by successively tying the measure of the Dirichlet prior to the statistics of lower orders.

### 3.2. Introducing Latent Topics

The graphical model for the Variable-gram Topic model is depicted in Figure 2(b). Similarly to LDA, each document is modelled as a mixture of latent topics and the latent topics are shared among documents. Each document $d$ has a distribution over the $K$ latent topics parametrised by $\boldsymbol{\theta}_d$, where $\boldsymbol{\theta}_d$ is defined as $\theta_{k|d} \equiv P(z = k|d)$. On the other hand, each topic is now represented by a Dirichlet-VMM; instead of a single probability distribution over words, we now have a set of conditional probability distributions encoding contextual information of variable order. This difference from LDA is apparent in Figure 2, where we can see that in the variable-gram topic model, word $w_t$ has directed connections from both $z_t$ and the $L$ previously observed words. These latter connections are defined in terms of the Dirichlet-VMM, which means that depending on the context we observe, we can have $\ell$ active connections, with $\ell \in \{1, \ldots, L\}$. If we consider only first order ($\ell = L = 1$) dependencies, instead of variable order ones, then we retrieve the Bigram Topic model of Wallach (2006).

**Algorithm 1** Generative Process for the Variable-gram Topic model.

**Input:** Dirichlet-VMM $\mathcal{T}$, $K$, $\boldsymbol{\Theta}$, $\boldsymbol{\Phi}$
**for** each document $d$ in the corpus $\boldsymbol{\omega}$ **do**
  **for** each time-step $t, t \in \{1, ..., T_d\}$ in $d$ **do**
    Choose a topic $z_{t,d} \sim \text{Multinomial}(\boldsymbol{\theta}_d)$
    Choose a word $w_{t,d} \sim \text{Multinomial}(\boldsymbol{\phi}_{\mathbf{c}_{w_{t,d}}, z_{t,d}})$
  **end for**
**end for**

Let $j$ index the leaf nodes of a Dirichlet-VMM, i.e. the contexts that can be used during prediction, and $\mathbf{c}_{w_t} = w_{t-1} \ldots w_{t-\ell}$, $\ell \in \{1, \ldots, L\}$ denote the context of word $w_t$. The parameters $\boldsymbol{\phi}_k$ characterising word generation within topic $k$ are defined by $\phi_{i|j,k} \equiv P(w_t = i|\mathbf{c}_{w_t} = j, z_t = k)$. This results in a tensor $\boldsymbol{\Phi}$ with $KC(W-1)$ free parameters, where $K$ is the number of latent topics, $C$ the number of leaf nodes and $W$ the number of words in the vocabulary. For simplicity we assume that the Dirichlet-VMM has the same tree structure for all topics.

According to the generative process in Alogrithm 1, the joint probability of a corpus $\boldsymbol{\omega}$ and a set of corresponding latent topic assignments $\mathbf{z}$ under a variable-gram topic model with parameters $\boldsymbol{\Phi}$ and $\boldsymbol{\Theta}$ is

$$P(\boldsymbol{\omega}, \mathbf{z}|\boldsymbol{\Phi}, \boldsymbol{\Theta}) = \prod_d \prod_t P(z_{t,d}|\boldsymbol{\theta}_d) P(w_{t,d}|\boldsymbol{\phi}_{\mathbf{c}_{w_{t,d}}, z_{t,d}})$$
$$= \prod_d \prod_i \prod_j \prod_k \phi_{i|j,k}^{N_{i|j,k}} \theta_{k|d}^{N_{k|d}} \ , \qquad (1)$$

where $N_{i|j,k}$ is the number of times word $i$ has been assigned to topic $k$ when preceded by context $j$, $N_{k|d}$ is the number of times topic $k$ has occured in document $d$ and $t$ indexes word positions (time-steps).

The prior over the $\boldsymbol{\Theta}$ parameters is the same as in LDA

$$P(\boldsymbol{\Theta}|\alpha\mathbf{n}) = \prod_d \text{Dirichlet}(\boldsymbol{\theta}_d|\alpha\mathbf{n}) \ . \qquad (2)$$

In this work, we set $\mathbf{n}$ to the uniform distribution.

The prior over the $\boldsymbol{\Phi}$ parameters is now defined in terms of the Dirichlet-VMM. More specifically, let $\mathbf{m}_{j,k} = \boldsymbol{\phi}_{\mathbf{pa}(j),k}$, denote the measure for the Dirichlet prior of node $j$ in the $k$-th Dirichlet-VMM. We have

$$P(\boldsymbol{\Phi}|\beta\{\mathbf{m}_{j,k}\}) = \prod_k \prod_j \text{Dirichlet}(\boldsymbol{\phi}_{j,k}|\beta\mathbf{m}_{j,k}) \ . \qquad (3)$$

This prior produces a hierarchical Dirichlet-VMM tree for each topic $k$. The parameters for the children of a node are coupled through the shared Dirichlet prior. Hence, within a topic, information regarding prediction is shared among all contexts, through the mutual prior at the root node. In this work we set the prior for the root node of each topic, $\mathbf{m}_{0,k}$, to a uniform distribution.

### 3.3. Inference & Learning

The total probability of the model given a set of hyperparameters is

$$P(\boldsymbol{\omega}, \mathbf{z}, \boldsymbol{\Phi}, \Theta|\alpha\mathbf{n}, \beta\{\mathbf{m}_{j,k}\}) = \prod_d P(\boldsymbol{\theta}_d|\alpha\mathbf{n})\times$$
$$\prod_{k,j} P(\boldsymbol{\phi}_{j,k}|\beta\mathbf{m}_{j,k}) \prod_t P(z_{t,d}|\boldsymbol{\theta}_d) P(w_{t,d}|\boldsymbol{\phi}_{\mathbf{c}_{w_{t,d}},z_{t,d}}). \quad (4)$$

From (4) we can see that, as with LDA, given $\alpha\mathbf{n}$, the $\boldsymbol{\theta}_d$ parameters are independent from each other and the same for all the $\boldsymbol{\phi}_{j,k}$. Similarly, the $\boldsymbol{\phi}_{j,k}$ parameters are independent from each other given $\beta$ and $\{\mathbf{m}_{j,k}\}$. Using these independence relations and the conjugacy of the Dirichlet to the multinomial, we can integrate over the model parameters, $\boldsymbol{\Theta}$ and $\boldsymbol{\Phi}$, to obtain a closed form solution for the joint probability of a corpus $\boldsymbol{\omega}$ and a set of corresponding latent topic assignments $\mathbf{z}$, given a set of hyperparemeters

$$P(\boldsymbol{\omega}, \mathbf{z}|\alpha\mathbf{n}, \beta\{\mathbf{m}_{j,k}\}) =$$
$$\prod_k \prod_j \frac{\Gamma(\beta)}{\prod_i \Gamma(\beta m_{i|j,k})} \frac{\prod_i \Gamma(N_{i|j,k} + \beta m_{i|j,k})}{\Gamma(N_{j,k}+\beta)} \times$$
$$\prod_d \frac{\Gamma(\alpha)}{\prod_k \Gamma(\alpha n_k)} \frac{\prod_k \Gamma(N_{k|d} + \alpha n_k)}{\Gamma(N_d + \alpha)} \quad . \quad (5)$$

Using (5) we can define a collapsed Gibbs sampling procedure that will allow us to infer the latent topic assignments. The procedure starts by randomly initialising the latent topic assignments, and then sequentially sampling each latent variable $z_t$ given the current values of all other latent variables $\mathbf{z}_{-t}$, the data $\boldsymbol{\omega}$ and a set of hyperparameters $\{\alpha\mathbf{n}, \beta\{\mathbf{m}_{j,k}\}\}$. At every Gibbs step we sample $z_t$ according to

$$P(z_t = k|\mathbf{z}_{-t}, \boldsymbol{\omega}, \alpha\mathbf{n}, \beta\{\mathbf{m}_{j,k}\}) \propto$$
$$\frac{\{N_{i|j,k}\}_{-t} + \beta m_{i,j,k}}{\{N_{j,k}\}_{-t} + \beta} \frac{\{N_{k|d}\}_{-t} + \alpha n_k}{\{N_d\}_{-t} + \alpha} \quad (6)$$

After the Gibbs sampling procedure has converged, we can approximate the posterior distribution of the model parameters, $\boldsymbol{\Phi}$ and $\boldsymbol{\Theta}$, through the predictive distributions

$$P(\theta_{k|d}|\boldsymbol{\omega}, \mathbf{z}, \alpha\mathbf{n}) = \frac{N_{k|d} + \alpha n_k}{N_d + \alpha} \quad (7)$$

$$P(\phi_{i|j,k}|\boldsymbol{\omega}, \mathbf{z}, \beta\{\mathbf{m}_{j,k}\}) = \frac{N_{i|j,k} + \beta m_{i|j,k}}{N_{j,k} + \beta} \quad (8)$$

## 4. Experiments

In the following section we evaluate the Variable-gram Topic model by comparing its performance with Latent Dirichlet Allocation, the Bigram Topic model and the Dirichlet-VMM. First, we consider a next-step prediction task, which is commonly used for evaluation in the music context (Lavrenko & Pickens, 2003; Paiement et al., 2009; Begleiter et al., 2004).

Although predictive log-likelihood is indicative of model performance, it only examines certain aspects of what a model has learnt. More specifically, log-likelihood decreases sharply if a model is overfitting, but it does not penalise as heavily a model that assigns lots of its probability mass to improbable configurations. This is problematic, as in unsupervised learning of complex data it is common for a model to underfit.

To address this issue, we introduce a novel framework for model evaluation which employs string kernels and the Maximum Mean Discrepancy (Gretton et al., 2006) to compare samples from the model with test sequences. This evaluation is indicative of underfitting, as models that spread their probability mass outside the space of possible configurations will generate samples that do not resemble data sequences. Therefore, this framework is complementary to the log-likelihood evaluation and allows us to further understand the generative properties of a model.

Finally, we provide a qualitative evaluation of the Variable-gram Topic model, where we analyse the inferred latent topic assignments and the learned parameters and show that the model captures musically meaningful properties, such as the key and the tempo.

### 4.1. Experimental Setup

For our experiments we use a dataset of MIDI files comprising 264 Sottish and Irish reels from the Nottingham Folk Music Database. Roughly half of the pieces are in the G major scale and the rest in the D major and all the pieces have 4/4 meter. We use the representation of Spiliopoulou & Storkey (2011), where time is discretized in eighth notes and the MIDI values are mapped to a 26-multinomial variable, with 24 values representing pitch (C4-B5) and 2 special values representing "silence" and "continuation". This representation is depicted in Figure 3.

To set the hyperparameters, $\alpha$ and $\beta$, of the topic models we use a 10-fold cross-validation procedure and perform grid search over the product space of the values $\{0.01, 1, 5, 10, 50, 100\}$.

We present results for topic models with 5, 10 and 50 topics. Additionally, for the Variable-gram Topic model and the Dirichlet-VMM, we present results using two tree structures, obtained by changing the threshold that the relative frequency of a context must exceed, in order to include the context in the tree. The first tree is relatively shallow (threshold: $1e-03$) and is referred to as $.Sh$, whereas the second tree is deeper (threshold: $1e-04$) and is referred to by $.De$.

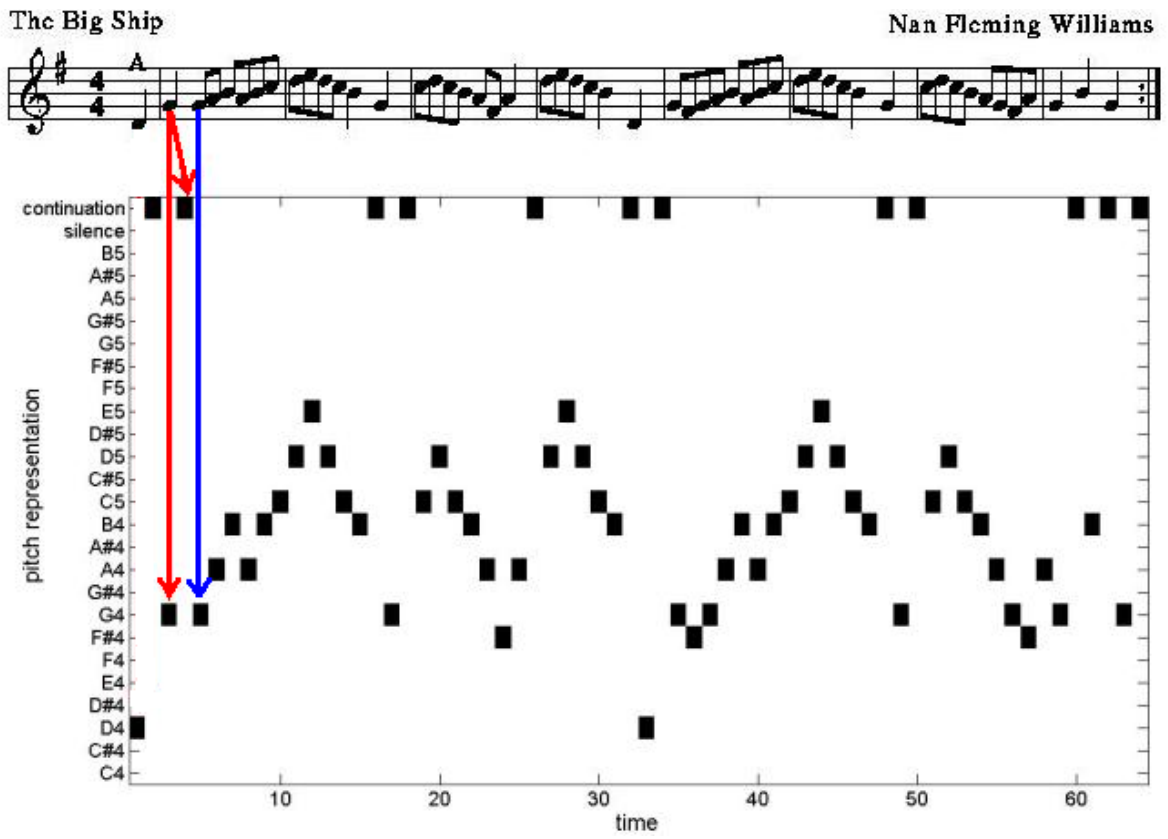

*Figure 3.* Data Representation: Time is discretized in eighth note intervals. At each time-step, pitch is represented by a 1-of-26 dimensional vector. Red arrows: a $G4$ quarter note (lasts for two eighths) is represented by $G4$ followed by 'continuation'. Blue Arrow: a $G4$ eighth note is represented by a single $G4$. Notes outside the $C4$–$B5$ interval are truncated to the nearest octave.

### 4.2. Next-Step Prediction Task

Given a test corpus of melodic sequences, we want to evaluate how well a model performs in next-step prediction. The average next-step prediction log-likelihood of a test corpus $\boldsymbol{\omega}_{test}$ under a model $M$ with parameters $\boldsymbol{\theta}$ is given by:

$$\mathcal{L} = \frac{1}{N}\sum_{d=1}^{N}\frac{1}{T_d}\sum_{t=1}^{T_d}\log P(w_{t,d}|w_{1,d},\ldots,w_{t-1,d}). \quad (9)$$

In the Dirichlet-VMM we can compute (9) exactly. For the topic models, computing the prediction log-likelihood requires a summation over the latent topic assignments for the test set, which is intractable. We approximate (9) through a sampling procedure, where we initialize $\boldsymbol{\theta}_d$ at the prior and at each time-step we sample $s$ topics from our current estimate of $\boldsymbol{\theta}_d$, use these samples to compute the log-likelihood of time-step $t$ and subsequently update $\boldsymbol{\theta}_d$ with the mean of the posterior distribution from each sample. Additionally we present results from two different update schemes for the distributions over words. In the first approach, denoted by S.1, we do not update the word distribution during testing, that is the information from the observed part of a test piece is only used to update the $\theta_d$ parameters. In the second approach, denoted by S.2, after each time-step we also update the distributions over words, by adding to $\phi_{i|j,k}$ of the observed word-context, counts proportional to the posterior probability of topic $k$.

*Table 1.* Average next-step prediction log-likelihood of the testdata under different models.

| Stationary Models | $\mathcal{L}(M, \boldsymbol{\theta}; \boldsymbol{\omega}_{test})$ | | |
|---|---|---|---|
| EmpMarg | −2.2216 | | |
| Dir-Bigram | −1.9153 | | |
| Dir-VMM.Sh | −1.6043 | | |
| Dir-VMM.De | −1.5605 | | |
| **Topic Models (S.1)** | $K=5$ | $K=10$ | $K=50$ |
| LDA | −2.0484 | −2.0355 | −2.0282 |
| Bigram | −1.7658 | −1.7558 | −1.7071 |
| Var-gram.Sh | −1.5680 | −1.5653 | −1.5349 |
| Var-gram.De | **-1.5390** | **-1.5427** | **-1.5219** |
| **Topic Models (S.2)** | $K=5$ | $K=10$ | $K=50$ |
| LDA | −2.0480 | −2.0351 | −2.0280 |
| Bigram | −1.7575 | −1.7433 | −1.6966 |
| Var-gram.Sh | −1.1724 | **-0.9827** | **-0.9553** |
| Var-gram.De | **-1.0354** | −0.9850 | −1.0194 |

Table 1 shows the average next-step prediction log-likelihood of the testdata under different models. Note that this is computed using only the first half of the test sequences, as in this genre the second half is typically exact repetition of the first half. The empirical marginal distribution, denoted as EmpMarg, is used as a simple baseline. It is the Maximum Likelihood model if we do not include any temporal dependencies or topic components. The Dir-Bigram model is a Dirichlet-VMM with $L = 1$ and is included as a second baseline that models first order dependencies. The first thing we can note is that introducing latent topics is useful for modelling melody, as all the topic models perform better than their non-topic counterparts and performance improves as we use more topics. This is true for both update schemes S.1 and S.2.

A second observation is that modelling temporal dependencies is very important in melody. The Dirichlet-VMM, which has no latent topics, but is able to capture both large and small order Markov dependencies, performs better than the Topic Bigram and LDA, which model only first order and no temporal dependencies respectively. Similarly, the Dirichlet-Bigram performs better than LDA. Therefore, the predictive information of the context is higher than that of the latent topics. This is also evident by the fact that performance improves as we consider longer contexts, with the Variable-gram topic models being consistently better than all other models for both update schemes S.1 and S.2.

Finally, the aspect of novelty in music is particularly apparent when we compare the performance of the Variable-gram topic models using update scheme S.1 and S.2. We can see that performance improves significantly when the distributions over notes are updated during prediction, which shows that information coming from the test piece is highly predictive of the future. This signifies that longer contexts can have different meaning across music pieces, which is expected as each piece is a unique realisation of a music idea.

### 4.3. Maximum Mean Discrepancy of String Kernels

In this section, we present a new approach for evaluating model generation. We employ string kernels and the Maximum Mean Discrepancy (Gretton et al., 2006) to estimate the distance between the model distribution, $Q$, and the true "theoretical" data distribution, $P$, based on finite samples drawn i.i.d. from each. Given the two populations – model samples and test data – we first compute a similarity score between each pair of sequences, which is proportional to the number of matching subsequences. Then we quantify the distance between the two populations by comparing the intra-population similarity scores to the inter-population scores. A small distance indicates that a model generates many of the different substructures that occur in the data. The method cannot assess the generalisation properties of a model, but it identifies underfitting, by measuring how close are model generations to data sequences. Therefore, it provides a complementary evaluation of model performance.

The Maximum Mean Discrepancy (MMD) is a distance metric between probability distribution embeddings on a Reproducing Kernel Hilbert Space (RKHS). Given a set of observed data sequences $\mathbf{X} := \{\mathbf{x}_1, \mathbf{x}_2, \ldots, \mathbf{x}_m\}$ drawn i.i.d. from $P$ and a set of sampled sequences $\mathbf{X}' := \{\mathbf{x}'_1, \mathbf{x}'_2, \ldots, \mathbf{x}'_n\}$ drawn i.i.d. from $Q$, an unbiased empirical estimate of the squared population MMD can be computed as

$$MMD_u^2\left[\mathcal{F}, \mathbf{X}, \mathbf{X}'\right] = \frac{1}{m(m\text{-}1)} \sum_{i=1}^{m} \sum_{j \neq i}^{m} K(\mathbf{x}_i, \mathbf{x}_j) + \frac{1}{n(n\text{-}1)} \sum_{i=1}^{n} \sum_{j \neq i}^{n} K(\mathbf{x}'_i, \mathbf{x}'_j) - \frac{2}{mn} \sum_{i=1}^{m} \sum_{j=1}^{n} K(\mathbf{x}_i, \mathbf{x}'_j), \quad (10)$$

where $\mathcal{F}$ is a RKHS and $K(x, x') = \langle \phi(x), \phi(x') \rangle_{\mathcal{F}}$ is a positive definite kernel defined as the inner product between feature mappings $\phi(x) \in \mathcal{F}$.

Since the MMD is defined on a RKHS, we can use the kernel trick to compare pairs of melodic sequences. String kernels naturally lend themselves to this problem, as they define a measure of similarity between discrete structures by comparing the set of matching substructures. We use the mismatch kernel (Leslie et al., 2004), $K_{(k,m)}(\mathbf{x}, \mathbf{x}')$, which for a pair of sequences $\mathbf{x}$ and $\mathbf{x}'$ computes the shared occurences of $k$-length subsequences that have at most $m$ mismatches. This kernel has been successfully used for biological sequence classification (Leslie et al., 2004) and NLP tasks (Teo & Vishwanathan, 2006).

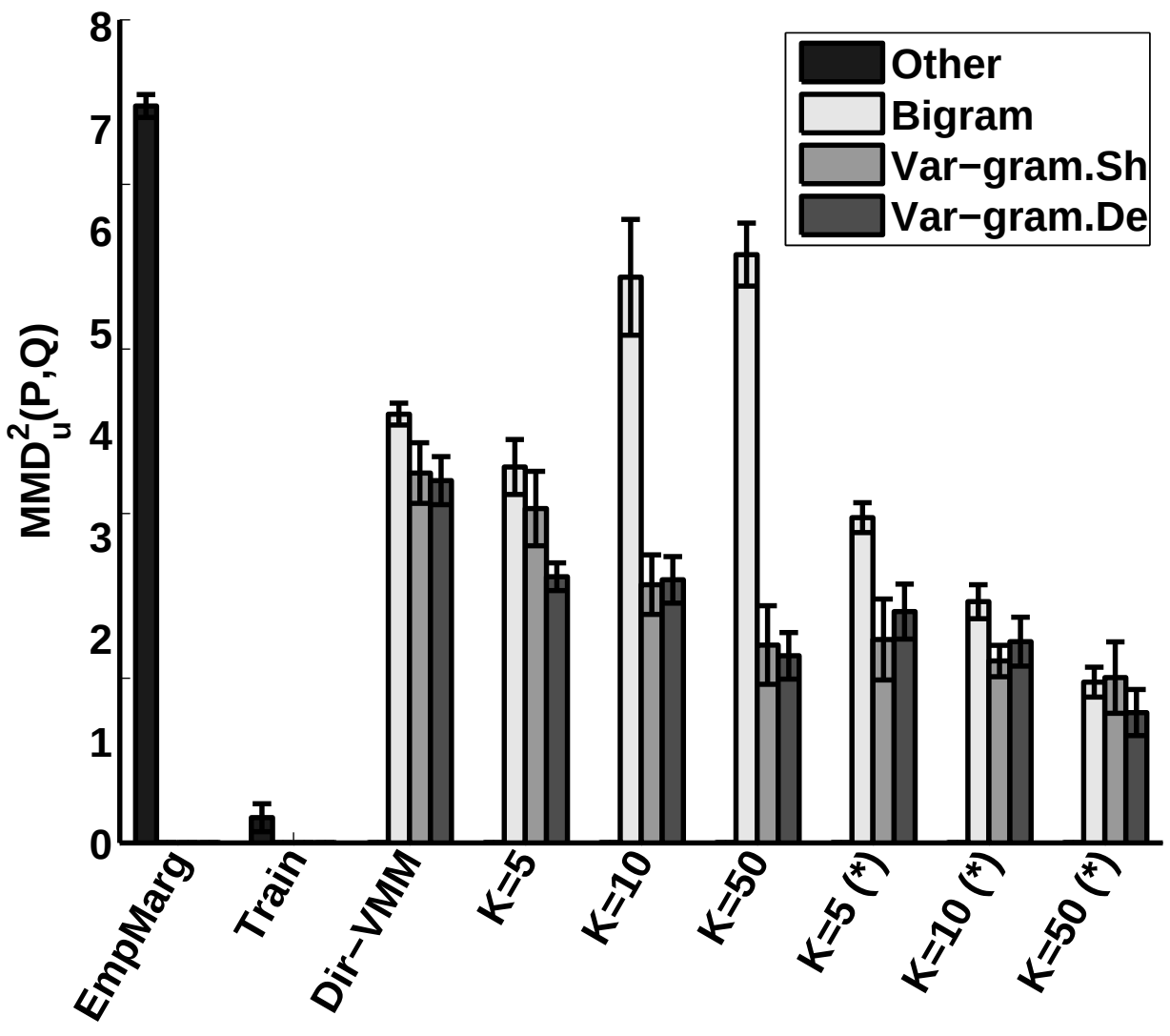

*Figure 4.* Estimated squared MMD between test sequences and model samples using the $(4, 1)$ mismatch kernel. Colour describes the temporal structure in a model. Groups describe models with different numbers of topics. EmpMarg: the empirical marginal distribution of the training data. Train: the train sequences. (*): The models are sampled conditional on the topic allocations (see text for details).

In order to avoid spurious correlations due to the "continuation" value, we map the melodic sequences to a 25-multinomial representation, where "continuation" is substituted by the observed pitch. Additionally, we report results using the normalized mismatch kernels

$$\tilde{K}_{(k,m)}(\mathbf{x}, \mathbf{x}') = \frac{K_{(k,m)}(\mathbf{x}, \mathbf{x}')}{\sqrt{K_{(k,m)}(\mathbf{x}, \mathbf{x})}\sqrt{K_{(k,m)}(\mathbf{x}', \mathbf{x}')}} \quad (11)$$

Figure 4 shows the mean and standard deviation of the estimated squared MMD between test sequences and model samples from different models, computed using the $(4, 1)$ mismatch kernel. Note that for the topic models we generate samples in two ways. In the first case, the topics are sampled from the prior which at each step is updated with the sampled topic. In the second case, which is denoted with the (*) symbol, we first run Gibbs sampling on the test sequences to get a set of topic allocations and then we perform sampling from the model given the topic allocations. This restricts the sampling noise, but is not directly comparable with the non-topic models, as it uses information from the test pieces captured by the latent topic allocations.

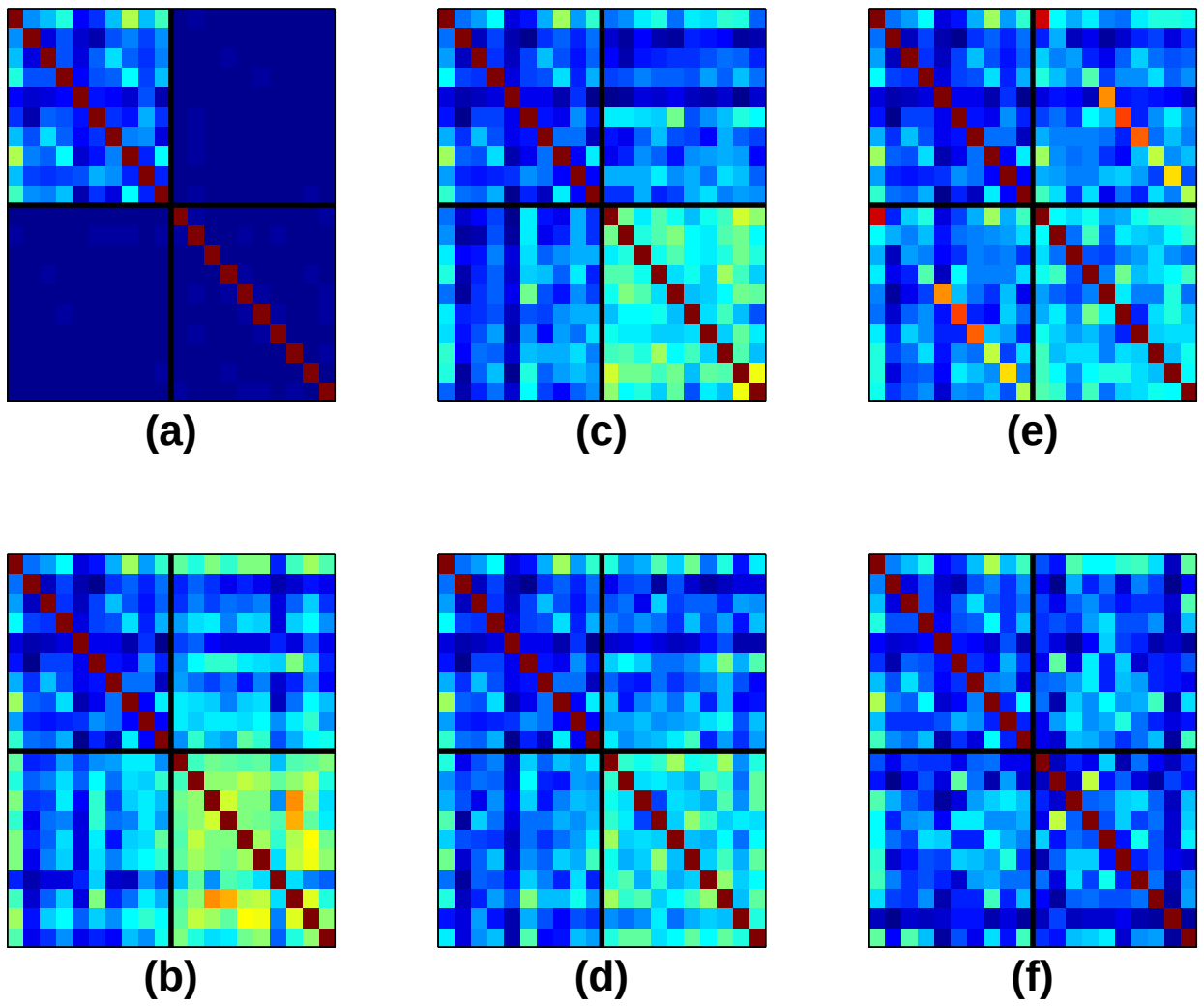

*Figure 5.* (4, 1) mismatch kernels computed between 10 test sequences (rows 1-10) and 10 model samples (rows 11-20). (a) Uniform(26) (b) Dirichlet-VMM.De (c) Topic Bigram.50 (d) Topic Var-gram.De.50 (e) Topic Var-gram.De.50-Given topics (f) Traindata

Again, we use the empirical marginal distribution as a baseline. Additionally, the MMD between test sequences and train sequences, denoted as Train, is given as a lower bound on what can be achieved. All models outperform the empirical marginal, but none of them has learned $P$ completely.

Although the objective function for this evaluation is very different to the one for the next-step prediction task, the comparative performance of the models under this metric is analogous. The first thing we can observe is that the topic component is useful. Similar to the results from the prediction task, the Variable-gram Topic model always outperforms the equivalent Dirichlet-VMM and performs better as we increase the number of topics. A second important observation is that introducing topics does not overcome the need for a systematic temporal model. The variable-gram models are consistently and notably better than the equivalent bigrams, and the variable-gram with the deeper tree structure (De) tends to outperform the shallower one (Sh). This indicates that a good model for temporal dependencies is important for both prediction and generation. Experiments using the $(5, 1)$ and $(6, 1)$ mismatch kernels produce equivalent results.

Figure 5 shows the $(4, 1)$ mismatch kernels between 10 test sequences and 10 samples from different models. We can observe that due to the lack of any temporal structure, samples from the Uniform distribution have very low similarity scores, both with each other and with test sequences. On the other hand, samples from the Dirichlet-VMM are comparatively much more similar to each other than to test sequences. Samples from the topic models are less similar to each other, as different samples have different distributions over topics, thus allowing for more unique generations, while still capturing the statistical dependencies across pieces. Another interesting observation is that when given the topic allocations for the test pieces (e), the samples have significantly more shared occurences with the corresponding test pieces, suggesting that the latent topic is highly informative of the observed note. This is depicted by the high values in the diagonal of the data-samples submatrix (top-right) in subfigure (e).

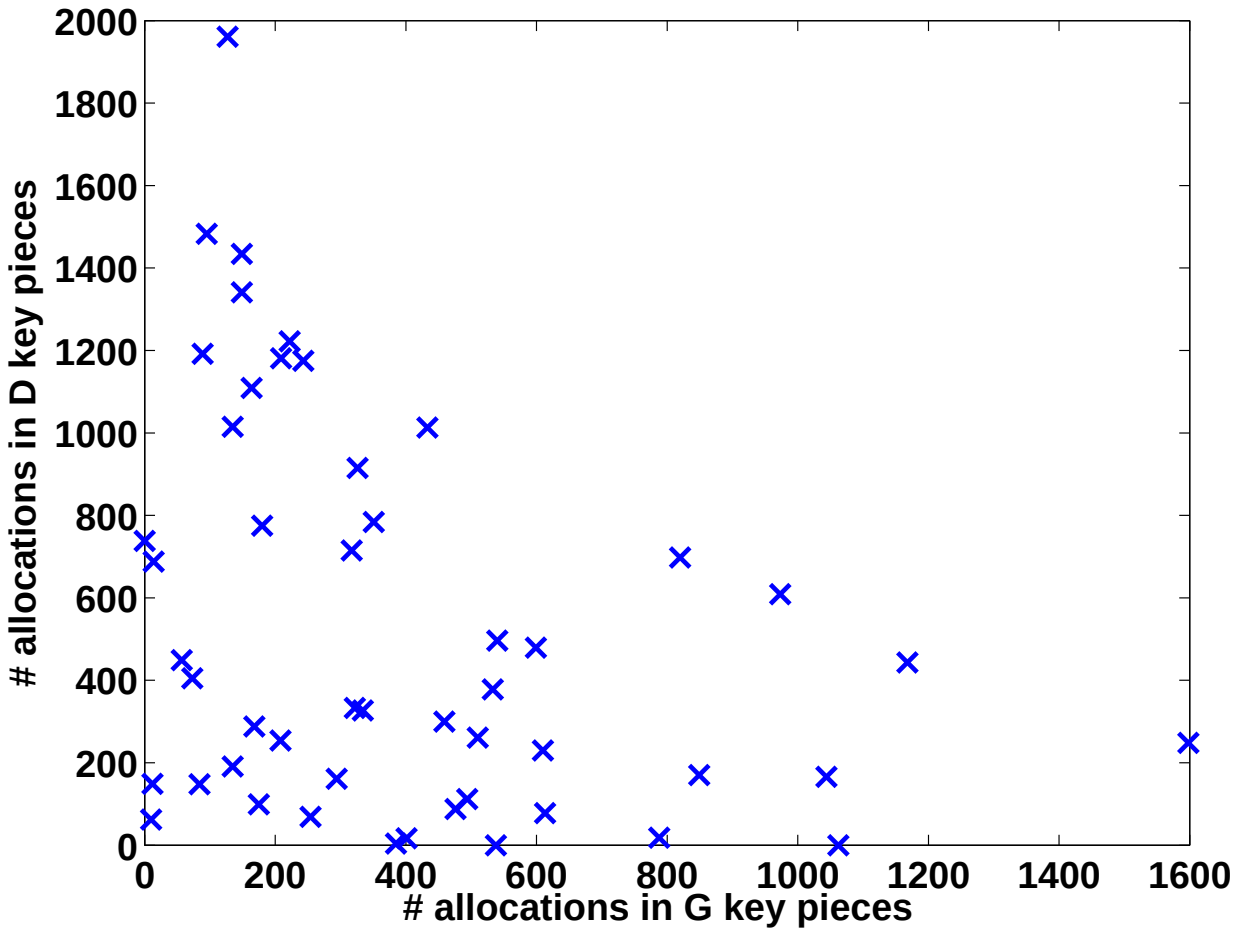

*Figure 6.* Scatter plot of the number of times each latent topic is allocated in pieces from the key of G (x-axis) and in pieces from the key of D (y-axis), after Gibbs sampling in the training data has converged.

## 4.4. Qualitative evaluation

An attractive property of Topic models when applied in NLP tasks is that they discover meaningful topics. In this section we examine aspects of the latent topic allocations and the inferred parameters of the Variable-gram topic model and analyse them with respect to musical features. Figure 6 shows a scatter plot of the number of times each topic has been allocated in pieces from the key of G (x-axis) and in pieces from the key of D (y-axis) after the Gibbs sampler has converged. The model has learned to distinguish the key, as topic allocations in this plot are negatively correlated, which means that each topic tends to be allocated in pieces from a single key. Figure 7 shows the conditional distributions over notes for 2 topics that are used in pieces from the key of D for 4 different contexts. The first context is the empty string, i.e. the distribution of the root node of the Dirichlet-VMM tree corresponding to each of the topics. The second context is note E5 and the other two are note E5 preceded by D5 and F#5 repsectively. We can see that the conditional distributions are successively sharper. Topic 4 tends to prefer "continuation", thus modelling slower parts of the melody, whereas topic 1 assigns lower probability to "continuation", especially given the longer contexts. At the same time we can observe that, depending on the context, the topics prefer different notes. For instance, if there is an upward movement in the context, D5 followed by E5 (subfigure (c)), Topic 4 wants to continue going upwards, i.e. high probability on F#5, and vice versa for a downward movement (subfigure (d)).

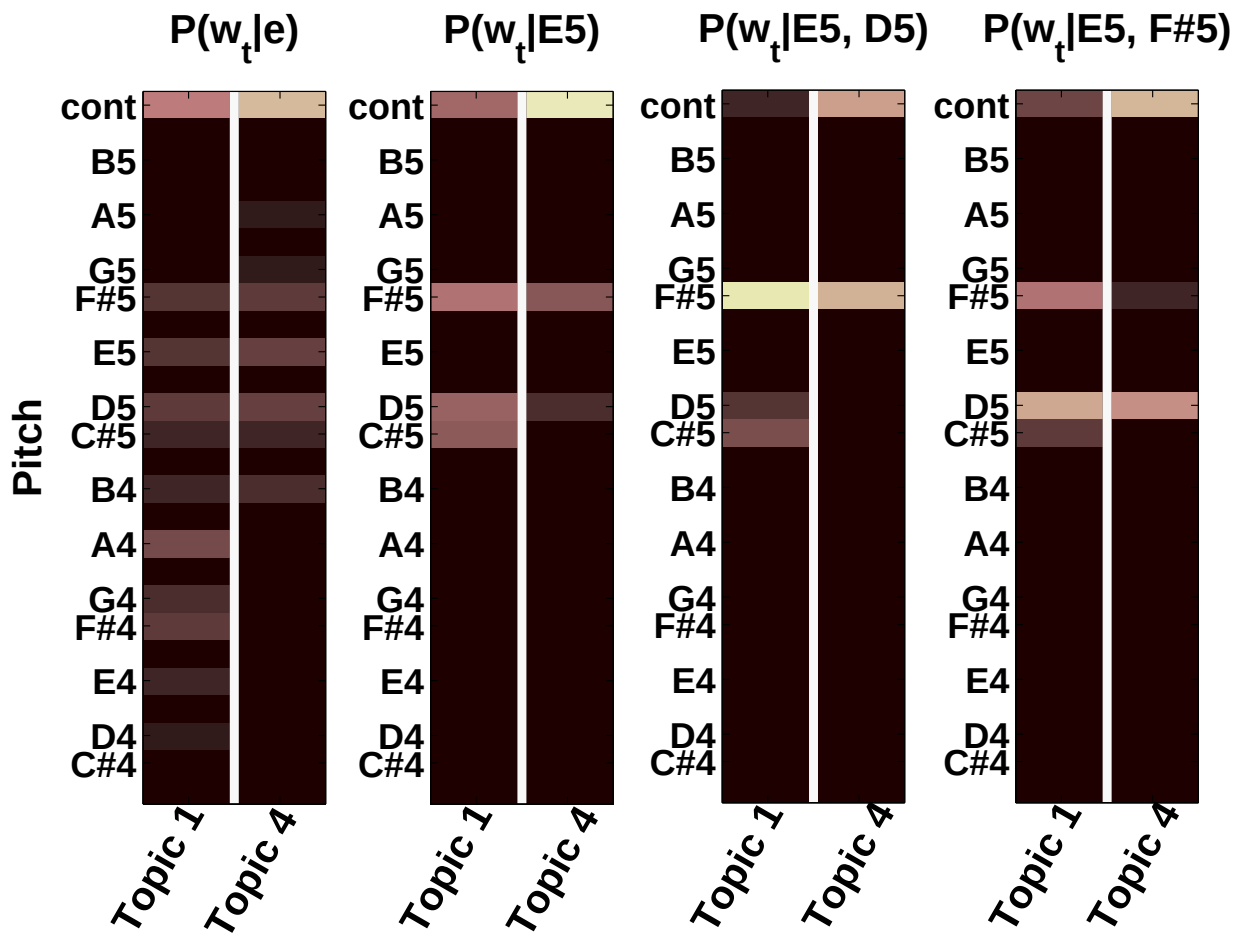

*Figure 7.* Conditional distributions over notes for 2 topics that are mostly assigned in pieces from the key of $D$. Each distribution is conditioned on a different context. Model: Variable-gram.De.$T_{50}$.

## 5. Discussion

We presented the Variable-gram Topic model, which couples the latent topic formalism with an expressive model of contextual information. Using two evaluation objectives we showed that the model outperforms a number of related methods for the problem of modelling melodic sequences. Both evaluations revealed that in this setting although latent topics improve performance, they do not overcome the need for a systematic temporal model.

Additionally we presented a novel way of evaluating model performance, where we used the MMD of string kernels computed between data sequences and model samples to directly evaluate the model distribution. This evaluation gave the same comparative results as next-step prediction, although it addresses different aspects of model behaviour. Looking at the mismatch kernels in Figure 5 we can visualize the progress that a model has made and "how much" of the structure is still not captured.

Finally it is interesting to investigate the aspect of novelty in music. Pearce & Wiggins (2004) moves in this direction by using the cross product of two models, one constructed using the train data and the other constructed sequentially as we observe the test sequence. Although this is applicable in a prediction task, it is not easy to actualize for generation, as the only available information comes from what the model has previously sampled.

### Acknowledgments

The authors would like to thank Iain Murray and the anonymous reviewers for providing useful comments.